\documentclass[draftclsnofoot, onecolumn, comsoc, 12pt]{IEEEtran} 
\usepackage{amsmath,amsfonts}
\usepackage{algorithmic}
\usepackage{algorithm}
\usepackage{array}
\usepackage[caption=false,font=normalsize,labelfont=sf,textfont=sf]{subfig}
\usepackage{textcomp}
\usepackage{stfloats}
\usepackage{url}
\usepackage{verbatim}
\usepackage{graphicx}
\usepackage{cite}
\usepackage{comment}
\usepackage{amsmath,amssymb,amsfonts,amsthm}
\usepackage{multirow}
\usepackage{colortbl}
\usepackage{diagbox}
\usepackage{color}
\usepackage{adjustbox}
\usepackage{float}
\usepackage{lipsum}
\usepackage[table]{xcolor}
\usepackage{subfig}
\usepackage{boldline}
\usepackage{hhline}
\usepackage{soul}
\usepackage[group-separator={,},group-minimum-digits=4]{siunitx}

\definecolor{forestgreen}{rgb}{0.13, 0.55, 0.13}
\definecolor{LightBlue}{rgb}{0.75,0.936,1.00}
\sethlcolor{LightBlue}
\definecolor{LightCyan}{rgb}{0.88,1,1}

\usepackage{wrapfig}
\usepackage{eucal}
\usepackage{pifont,bbding,fontawesome}

\def\BibTeX{{\rm B\kern-.05em{\sc i\kern-.025em b}\kern-.08em
    T\kern-.1667em\lower.7ex\hbox{E}\kern-.125emX}}
\definecolor{abstractbg}{rgb}{0.89804,0.94510,0.83137}
\begin{document}

\title{PC-DeepNet: A GNSS Positioning Error Minimization Framework Using Permutation-Invariant Deep Neural Network}

\author{M. Humayun Kabir, Md. Ali Hasan, Md. Shafiqul Islam, \\Kyeongjun Ko,~\IEEEmembership{Member,~IEEE}, and Wonjae Shin,~\IEEEmembership{Senior Member,~IEEE}
%
\thanks{M. Humayun Kabir is with the Department of Electrical and Electronic Engineering, Islamic University, Kushtia 7003, Bangladesh (e-mail: \mbox{humayun@eee.iu.ac.bd}).} \thanks{Md. Ali Hasan and Wonjae Shin are with the School of Electrical Engineering, Korea University, Seoul 02841, Republic of Korea (e-mail: alihasan@korea.ac.kr; wjshin@korea.ac.kr).} \thanks{Md. Shafiqul Islam is with the Department of Computer Science and Engineering, Bangladesh University of Business and Technology, Dhaka 1216, Bangladesh (e-mail: \mbox{msislam@bubt.edu.bd}).} \thanks{Kyeongjun Ko is with the School of Electronics and Electrical Engineering, Hongik University, Seoul 04066, Republic of Korea (e-mail: kkj8000@hongik.ac.kr).}
%
}

\maketitle


\begin{abstract}
Global navigation satellite systems (GNSS) face significant challenges in urban and sub-urban areas due to non-line-of-sight (NLOS) propagation, multipath effects, and low received power levels, resulting in highly non-linear and non-Gaussian measurement error distributions.
In light of this, conventional model-based positioning approaches, which rely on Gaussian error approximations, struggle to achieve precise localization under these conditions. To overcome these challenges, we put forth a novel learning-based framework, \textit{PC-DeepNet}, that employs 
a permutation-invariant (PI) deep neural network (DNN) to estimate position corrections (PC). This approach is designed to ensure robustness against changes in the number and/or order of visible satellite measurements, a common issue in GNSS systems, while leveraging NLOS and multipath indicators as features to enhance positioning accuracy in challenging urban and sub-urban environments. To validate the performance of the proposed framework, we compare the positioning error with state-of-the-art {model-based and learning-based} positioning {methods} using two publicly available datasets. 
The results confirm that proposed \mbox{PC-DeepNet} achieves superior accuracy than existing {model-based and learning-based} methods while exhibiting lower computational complexity compared to previous learning-based approaches.

\end{abstract}


\section{Introduction}
\label{sec:introduction}
\IEEEPARstart{I}{nternet} of Things (IoT) requires ubiquitous positioning to locate things everywhere. The convergence of Global Navigation Satellite Systems (GNSS) and IoT has paved a new era of location-based intelligence applications (e.g., asset tracking, precision agriculture, smart cities and infrastructure management, and safety and emergency response).  
GNSS provides a widely applicable satellite-based global positioning system, offering accurate and reliable geolocation data in the global reference frame. This capability supports the diverse needs of IoT applications across various domains, enabling a new era of efficiency, safety, and innovation. By leveraging precise location information, GNSS facilitates accurate decision-making and operational optimization. Several global and regional navigation systems provide positioning services under the umbrella of GNSS: Global positioning system (GPS), Galileo, BeiDou, and GLONASS in GNSS are used for providing service globally \cite{kabir2022performance}. In contrast, NavIC and Quasi-Zenith Satellite System (QZSS) are used for regional coverage. 
 However, due to the long orbital altitude of GNSS satellites, the signal strength at the Earth's surface is very weak.

The potential sources of error in GNSS include satellite clocks, orbit errors, ionospheric delays, tropospheric delays, receiver noise, and multipath. The typical magnitudes of these errors are as follows: $\pm2$ m for satellite clocks, $\pm2.5$ m for orbit errors, $\pm5$ m for ionospheric delays, $\pm0.5$ m for tropospheric delays, and $\pm0.3$ m for receiver noise \cite{li2023towards}. To mitigate ionospheric error and tropospheric error, the Klobuchar model \cite{klobuchar2007ionospheric} and the Saastamoinen model \cite{saastamoinen1972atmospheric} are commonly employed, respectively.
 The relative geometry between the user and the satellite also significantly affects the signal quality. Moreover, signals experience attenuation and reflection due to natural obstructions (e.g., large hills, trees) and man-made structures (e.g., buildings), leading to non-line-of-sight (NLOS) and multipath effects. Therefore, the measurement error distribution does not strictly follow the Gaussian distribution \cite{wang2013research}. The multipath signal is a delayed and attenuated replica of the line-of-sight (LOS) signal. It causes the receiver tracking loop to create some synchronization bias with respect to the LOS signal that induces errors in pseudorange measurements. As a result, positioning errors can exceed $\num{10}$ m \cite{ozeki2022gnss}. Since GNSS enables real-time tracking of people, assets, and objects, precise location is crucial in IoT applications.

There are several approaches to {minimize} the effect of pseudorange measurement uncertainty to reduce positioning errors: signal processing \cite{van1994multipath,townsend1994practical} antenna design \cite{jiang2014nlos,tranquilla1994analysis} modeling in the measurement domain, and positioning domain. In addition, advancements in machine learning (ML) techniques show promise in further enhancing the effectiveness of measurement domain modeling for multipath and NLOS error mitigation \cite{phan2013unified,yozevitch2016robust,socharoentum2016machine,quan2018convolutional,sun2019gps,sun2020improving}.
The measurement domain approaches detect multipath and classify LOS/NLOS signals. Since the measurement domain method utilizes LOS signals excluding NLOS in position estimation that cannot ensure good geometry and the number of LOS signal measurements can be insufficient to estimate the position in the urban environment.
In contrast, the positioning domain approach imposes a correction to an initial guess position utilizing different features extracted from GNSS measurements \cite{kanhere2022improving}.

The measurement error distribution is highly non-linear and non-Gaussian in urban and sub-urban areas due to the NLOS and multipath effects. Bearing in mind the conventional model-based positioning approach based on Gaussian approximations of the error could not handle such a non-linear and urban high-obstruction environment, which induces significant errors in positioning. The model-free learning-based deep neural network (DNN) in the position domain can be an alternative solution. Unlike conventional models, DNN is capable of building an error model by learning the relation between different parameters of measurements with positioning errors from the environment. However, the number and/or order of the measurement data keeps changing in time, which can not be adjusted with the conventional DNN model as it follows fixed-order inputs.

Incorporating permutation invariant into the DNN model ensures that the model's output remains consistent irrespective of the order of inputs. Permutation-invariant models, which leverage symmetric functions such as sum or mean, facilitate faster convergence with invariance by allowing the model to concentrate on the underlying patterns or relationships in the data rather than learning the order-specific information. By reducing the complexity of the search spaces, these models can achieve faster convergence.

A set transformer-based DNN model is proposed in \cite{kanhere2022improving} to estimate the position correction from GNSS measurement. This method follows a linear layer with the ReLU activation function in the DNN to map position errors using LOS vector and pseudorange residuals. It does not consider any NLOS and multipath indicators, like carrier-to-noise ratio ($C/N_0$), and elevation angle \cite{phan2013unified,yozevitch2016robust,socharoentum2016machine} for the position correction model. For training the model, noises are added from a uniform distribution within the predefined range with ground truth position to guess the initial position. The amount of position correction is determined based on this initial guess position. This approach does not train the model with proper position corrections since it considers only the predefined range of noises.

To address these shortcomings, we utilize a permutation-invariant (PI) DNN (PI-DNN) model consisting of an encoder, aggregation, and decoder. The aggregation layer utilizes the PI objective function (sum pooling) to ensure that the output remains fixed irrespective of the change in the order of inputs. 
{In comparison to traditional ReLU activation, Leaky ReLU helps to mitigate the risk of information loss associated with negative inputs.} Using the Leaky ReLU activation function in the PI-DNN model that maps data from some negative to positive infinity values. This approach increases the accuracy in positioning, ensuring that all the neurons in the network contribute to the output, even if their inputs are negative. 
We utilize $C/N_0$, satellite elevation angle, and geometrical dilution of precision (GDOP) as features with pseudorange residuals and LOS vector along $x, y,$ and $z$ coordinates for mapping with 3D-positioning error. The initial position estimates are obtained using robust weighted least squares (r-WLS) \cite{robust_WLS} method. These estimates are then refined by learning position corrections with respect to ground truth data. Specifically, we propose a framework named \textit{PC-DeepNet}, using the PI-DNN model to mitigate the positioning error in NLOS and multipath scenarios. For validation, we use two publicly available Android Raw GNSS measurement datasets \cite{fu2020android,fu2022android} which are collected from multiple driving paths in urban and suburban areas in San Francisco Bay and Los Angeles.

To summarize, the main contributions of this study are as follows:
\begin{enumerate}
\item We propose a PC-DeepNet framework using the PI-DNN model to handle the variation in the number and order of satellite measurements and minimize the positioning error. 
\item We introduce a feature extraction module that extracts seven different features from GNSS measurement: pseudorange residual, LOS vector, GDOP, $C/N_0$, and elevation angle for mapping with 3D-positioning error. Moreover, the position corrections are determined from the positioning difference of the initial guess position with respect to ground truth. 
\item We validate the proposed method and compare it with the existing state-of-the-art {model-based} methods {(WLS \cite{morton2021position}, r-WLS \cite{robust_WLS}, and Kalman Filter (KF) \cite{sever2022gnss}) and model-free learning-based method (Kanhere \textit{et al.} \cite{kanhere2022improving})} using the publicly available two datasets and achieve higher positioning accuracy {with low computational complexity} in the case of the proposed method. 
\end{enumerate}
The remainder of this study is summarized as follows: Section~\ref{sec: Related_Works} presents a detailed review of the existing technology based on satellite-based positioning. Section~\ref{sec: System Model} discusses the details of the system model. Section~\ref{sec: Dataset} presents the information about datasets.
Section~\ref{sec: Result_and_Discussion} describes the experimental results. Finally, Section~\ref{sec: Conclusion} concludes the study and discusses future work.

\section{Related Works}
\label{sec: Related_Works}
The measurement domain modeling-based methods use several ways to mitigate pseudorange measurement uncertainty.
The Bayesian estimators are widely used to mitigate the degradation of satellite signals. However, the KF \cite{sever2022gnss}, one of the most popular Bayesian estimators, is restricted by linearity and normal distribution assumptions that require explicit modeling of the effects to be mitigated. The dual-frequency observations and code-minus-carrier measurements in estimating the multipath condition are utilized in \cite{blanco2011multipath}. 
{This method can not mitigate the multipath error when the receiver is in motion.} Multipath and NLOS effects depend on the surrounding environment. Hence, geometric modeling of the degraded signals is useful in this regard. Analyzing the signal propagation path, possible reflection, and diffraction can be visualized with a 3D map. However, this method is computationally expensive, and its accuracy depends on the quality of the map \cite{obst2012urban}.

Sensor fusion techniques have also been introduced in several research studies to enhance positioning errors, and an inertial navigation system (INS) is used in this regard. 
INS is less induced on environmental conditions and provides linear acceleration and velocity measurements at a high output frequency.
Extended Kalman filter (EKF) based GNSS-INS system is effective in open areas with clear sky visibility. 
Moreover, the EKF-based system induces error in linearization steps. To address this issue, an iterative KF (IKF) is proposed in \cite{bell1993iterated} where multiple iterations are used in updating steps to prohibit the error generated in linearization steps.
{Both the EKF and IKF}-based system achieved the optimal estimation when the first-order Markov chain is considered, and random noise is in Gaussian distribution \cite{barfoot2017state,valiente2014comparison}. However, GNSS measurements can be non-Gaussian and highly time-correlated in dense urban areas \cite{wen2019tightly}. 
As a result, EKF {and IKF}-based sensor fusion fails to obtain optimal results in this area. As a remedy, a multi-state constrained Kalman Filter (MSCKF) \cite{li2013optimization} is {utilized that} considers geometric constraints of the feature measurements to update the states. 
However, the feature states are eliminated from the MSCKF using the nullspace matrix to reduce the size of the states. Specifically, the MSCKF does not fully utilize all the historical information, which affects the positioning accuracy \cite{wen2020time}. 
The fifth-generation (5G) network can be integrated with the global navigation satellite system (GNSS) to achieve highly accurate positioning. The 5G base stations (BSs) can provide range and angle measurements at a much higher rate than GNSS. A multiple-rate adaptive Kalman filter (MRAKF) for GNSS-5G hybrid positioning with a hybrid sequential fusing scheme is proposed in \cite{9721906}. This approach effectively integrates GNSS and 5G measurements at different data rates, leveraging the high-rate 5G measurements for enhanced accuracy. Experimental results demonstrate that the proposed MRAKF method significantly improves the positioning accuracy compared with the other adaptive noise estimation methods and the standard EKF. However, the user terminal (UT) equipment requires additional infrastructure and a direct communication link between the base station and UT, which may not always be feasible in IoT networks.

Several research works focus on applying ML for modeling and classifying the received signal into LOS and NLOS. Two complementary variables, satellite elevation, and azimuth angle, with a support vector machine (SVM), are utilized to classify the signal in \cite{phan2013unified}. Alternatively, a robust classifier adopting a decision tree (DT) with the $C/N_0$, elevation angle, and pseudorange as an input-based approach is proposed in \cite{yozevitch2016robust}. Four ML methods: logistic regression, SVM, Naive Bayes, and DT with a number of feature inputs, including satellite visibility, position dilution of precision (PDOP), and pseudorange corrections are utilized to detect NLOS signals in \cite{socharoentum2016machine}. Quan \textit{et al.} \cite{quan2018convolutional} introduce a convolutional neural network (CNN) with sparse autoencoder (SAE) to detect multipath. Sun \textit{et al.} \cite{sun2019gps} utilize several different complementary variables that affect the measurement error with principal component analysis (PCA) and artificial neuro-fuzzy inference system (ANFIS) to classify signals into LOS and NLOS.
For the above research work, it is assumed that most of the ML-based approach deals with the detection and classification of LOS and NLOS signals. A few works focus on applying ML as an estimator for the position error.
In \cite{sun2020improving}, authors proposed a gradient boosting decision tree (GBDT) based ML to improve positioning accuracy. It follows a two-way approach, positioning with pseudorange correction and multipath/NLOS signal exclusion or correction. They claim an improvement of position accuracy from $\num{81.3}$ m to $\num{23.3}$ m compared to the conventional method \cite{sun2019combining} {which does not satisfy user requirement.} Kanhere \textit{et al.} \cite{kanhere2022improving} propose a DNN with a set transformer that processes set-valued inputs derived from GNSS measurements to estimate the position correction. They utilize a linear layer with the ReLU activation function in the DNN and consider LOS vector and pseudorange residuals as features. For training, validating, and testing the proposed model, they consider the initial position guess with randomly sampled noise added to the true position. 
However, they do not consider any NLOS and multipath indicators such as $C/N_0$ and satellite elevation angle {\cite{phan2013unified,yozevitch2016robust,socharoentum2016machine}} as features for the position correction model. 
For training the model, noises are added from a uniform distribution within the predefined range with ground truth position to guess the initial position. The amount of position correction is determined based on this initial guess position. This approach does not train the model with proper position corrections since it considers only the predefined range of noises. 
Moreover, they present computational complexity in terms of model parameters of $\num{151107}$ and the required memory of the model is $\num{611}$~Kilobyte.
The computational complexity should be less to meet the hardware resources in practical applications.
To address these shortcomings, 
there are scopes to apply new ML solutions to build an effective error model by learning the relation between measurements with the positioning errors in NLOS and the multipath environment considering $C/N_0$ and satellite elevation angle including pseudorange residuals, LOS vector, and GDOP for the error model.

This paper proposes a {learning-based} PC-DeepNet framework that performs in three main modules: 1) Preprocessing, 2) PI-DNN Model, and 3) Corrected Receiver Position. PI-DNN model consists of three layers: i. Encoder, ii. Aggregation, and iii. Decoder. The preprocessing layer processes the GNSS measurement, extracts a feature vector with seven features, and sends it to the Encoder layer for encoding. The Encoder layer is a neural network (NN) consisting of four hidden layers with a Leaky ReLU activation function and $\num{2}\%$ dropout. In the Aggregation layer, the sum-pooling is used as a permutation invariant function. Finally, the Decoder decodes the encoded input using an NN with four hidden layers along with a Leaky ReLU activation function and a $\num{2}\%$ dropout. Moreover, to get the position correction with the ground truth, the positioning outcome of the pseudorange-based r-WLS estimation is considered as the initial guess position. 
The robust weighted least squares (r-WLS), combined with smooth L1 loss function, effectively reduces the influence of potential outliers in data. Smooth L1 loss is a variant of the standard L1 loss that mitigates sensitivity to outliers by smoothly transitioning from a quadratic function for small errors to a linear function for large errors.
Therefore, the proposed \mbox{PC-DeepNet} with seven features achieved a remarkable performance compared to the existing state-of-the-art methods in terms of positioning accuracy and computational complexity.

\section{System Model}
\label{sec: System Model}
\begin{figure*}
    \centering
    \includegraphics[width=\linewidth]{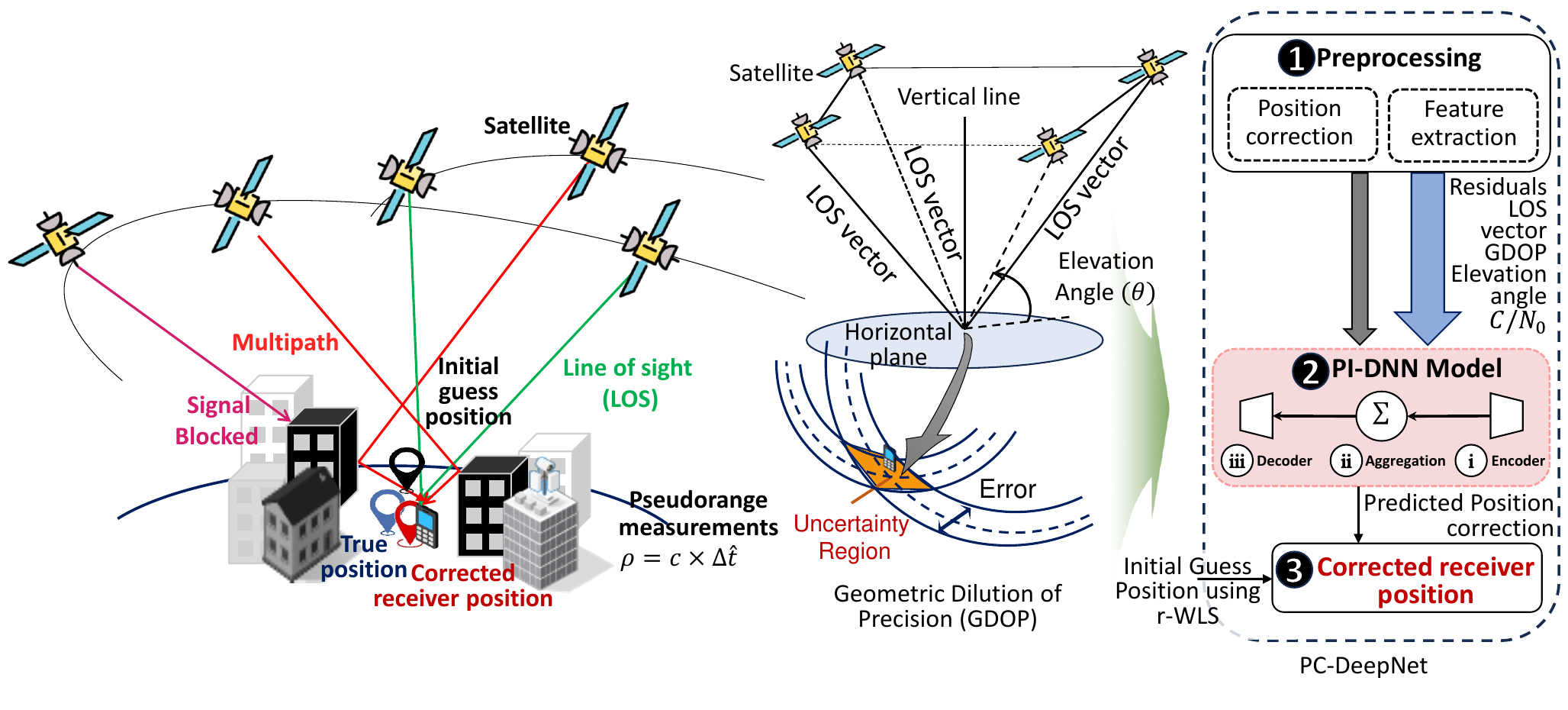}
    \caption{{The proposed PC-DeepNet to minimize the positioning error in NLOS and multipath environment.}}
    \label{system_model}
\end{figure*}

The GNSS combined several medium Earth orbit (MEO) and geostationary Earth orbit (GEO) constellations under an umbrella to provide outdoor positioning services.
The signal strength is very low since the distance between the satellites and the UT is too large. In addition, attenuation and reflection occur mostly in urban and sub-urban areas due to tall natural and man-made structures, which cause NLOS and multipath signals. In case of heavy obstacles, the signal may also be blocked. Multipath creates some synchronization bias with respect to the LOS signal, and as a result, it induces errors in pseudorange measurements. Therefore, the positioning outcome becomes inaccurate, which needs to be corrected. Pseudorange residuals, along with the LOS vector, indicate the amount of correction that needs to be added with the initial guess position. The relative geometry represented by the metric GDOP indicates the quality of positioning. Moreover, signals from the satellites with low elevation angles are prone to multipath effects. The $C⁄N_0$ of the satellite can be an indicator for the detection of NLOS and multipath signals. It is challenging to model NLOS and multipath error models analytically. 
The DNN model can be explored in this case as it learns the relationship between measurements and outcomes. All these features (pseudorange residuals, LOS vector, GDOP, elevation angle, $C⁄N_0$) can be used to map with the position error using suitable DNN architectures. The number and order of satellite measurements change over time, and it does not restore any impact on the positioning.
In conventional DNN, the number and order of inputs should always be fixed. 
It fails to adapt to dynamic environments when the number and order of the input change.
Whereas {the utilization of a PI-DNN model, which incorporates a PI objective function, ensures robustness/ resilience against changes in the number and order of satellite measurements over time, making it a highly suitable solution for addressing the dynamic nature of satellite-based positioning systems.} The positioning outcome of the r-WLS is taken as the initial position guess while determining the positioning correction. Fig.~\ref{system_model} shows the system architecture of the proposed \mbox{PC-DeepNet} to minimize the positioning error in the case of NLOS and multipath environments.

\label{sec: Proposed_methods}
\subsection{Preprocessing}
\label{sec: preprocess}
\begin{figure*}[h]
    \centering
    \includegraphics[width=\linewidth]{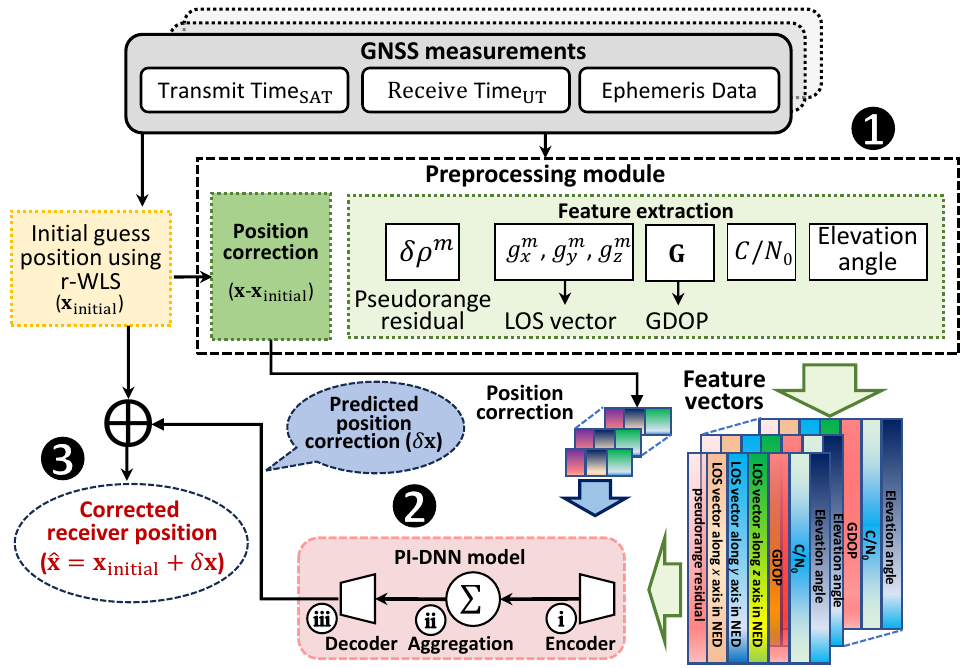}
    \caption{Proposed framework (PC-DeepNet) for positioning error minimization.}
    \label{dnn_framework}
\end{figure*}

The preprocessing section is responsible for extracting features and computing position correction for a certain time instance.
In preprocessing, we extract seven features {(Pseudorange residual, LOS vector, GDOP, $C⁄N_0$, and elevation angle)} and determine corresponding position correction using a number of pseudorange measurements collected from the visible satellites with respect to the initial guess user position ($\mathbf{x}_{\sf initial}$). Fig.~\ref{dnn_framework} shows the proposed \mbox{PC-DeepNet} architecture that minimizes the positioning error in NLOS and multipath environments.

\subsubsection{Pseudorange residual}
\label{sec: Pr}
The pseudorange residual provides the difference between the measured pseudorange and the estimated pseudorange. It indicates the amount of estimated error. The pseudorange measurement is defined as follows
\begin{equation} 
\rho= {\sf c}\times \Delta \hat{t},\label{eq1}
\end{equation}
where ${\sf c}$ = {speed} of radio frequency (RF), $\Delta \hat{t}$= estimated propagation time of the signal from the satellite to the receiver.

The pseudorange measurement for the $m$-th satellite is defined as follows \cite{noureldin2012fundamentals}
\begin{equation} 
\rho^m= r^m +{\sf c} \delta t_{\sf r} -{\sf c} \delta t_{\sf s} + {\sf c}I^m + {\sf c}T^m + \epsilon_{\rho}^m,\label{eq2}
\end{equation}
where $r^m$ is the true range between the receiver antenna at time $t_{\sf r}$ and the satellite antenna at time $t_{\sf t}$ (meters), $\delta t_{\sf r}$ is the receiver’s clock offset (sec), $\delta t_{\sf s}$ is the satellite’s clock offset (sec), $I^m$ is the ionospheric delay (sec), $T^m$ is the tropospheric delay (sec), and $\epsilon_{\rho}^m$ is the error in the range due to various sources, including receiver noise, multipath, and orbit prediction (meters).

After compensating for the satellite clock bias, ionospheric errors, and tropospheric errors, the corrected pseudorange yields
\begin{equation} \rho^m= r^m +b_{\sf r} + \tilde{\epsilon}_{\rho}^m,\label{eq-3}\end{equation}
where $b_{\sf r}={\sf c} \delta t_{\sf r}$ is the error in range (in meters) due to the receiver’s clock bias and $\tilde{\epsilon}_{\rho}^m$ is the total effect of residual errors.

The geometric range from the $m$-th satellite to the receiver is
\begin{eqnarray}
  r^m & = &\sqrt{(x-x^m)^2+(y-y^m)^2+(z-z^m)^2} \nonumber \\ 
&\approx & \left\|\mathbf x -\mathbf{x}^m\right\| ,\label{eq-4}  
\end{eqnarray} 
where $\mathbf x=[x,y,z]^T$ is the true receiver position in the Earth center Earth fixed (ECEF) frame, and
$\mathbf{x}^m=[x^m,y^m,z^m ]^T$ is the position of the $m$-th satellite in ECEF frame.

Substituting \eqref{eq-4} in \eqref{eq-3} we can get
\begin{equation} \rho^m= \left\|\mathbf x-\mathbf{x}^m\right\| +b_{\sf r} + \tilde{\epsilon}_{\rho}^m .\label{eq3}\end{equation}

Linearizing around the initial guess receiver position $\mathbf{x}_{\sf initial}=[x_{\sf initial},y_{\sf initial},z_{\sf initial} ]^T$ using first-order Taylor series with respect to $\mathbf{x}$, we can get \eqref{eq-6}

\begin{figure*}
\begin{align}   
\hline \notag \\ 
\rho^m &= \sqrt{\left (x_{\sf initial}-x^m \right)^2+\left(y_{\sf initial}-y^m\right)^2+\left(z_{\sf initial}-z^m\right)^2} \nonumber\\ &\quad+\frac{\left(x_{\sf initial} -x^m\right)\left(x -x_{\sf initial}\right) + \left(y_{\sf initial} -y^m\right)\left(y -y_{\sf initial}\right)+ \left(z_{\sf initial} -z^m\right)\left(z -z_{\sf initial}\right)}{\sqrt{\left(x_{\sf initial}-x^m\right)^2+\left(y_{\sf initial}-y^m\right)^2+\left(z_{\sf initial}-z^m\right)^2}} +b_r + \tilde{\epsilon}_{\rho}^m  \label{eq-6}\\ 
 \hline \notag 
\end{align} 
\end{figure*}

The pseudorange residual can be denoted as
\begin{equation} \delta\rho^{m}= \mathbf{g}_{\sf initial}^m \cdot \delta \mathbf{x} + \delta b_{\sf r} + \tilde{\epsilon}_{\rho}^m,\label{eq4}\end{equation}
where $\delta \rho^m=\rho^m- \rho_{\sf initial}^m $, $\rho_{\sf initial}^m$ = estimated pseudorange between satellite $m$ and initial guess receiver position $(\mathbf{x}_{\sf initial})$, $(\mathbf{g}_{\sf initial}^m)^T=[g_x^m,g_y^m,g_z^m ]$, $\delta \mathbf{x}=\mathbf{x}-\mathbf{x}_{\sf initial}=[x,y,z]^T-[x_{\sf initial},y_{\sf initial},z_{\sf initial}]^T$, and $\delta b_{\sf r} = b_{\sf r} - b_{\sf r,initial}$.

The LOS vector along $x, y,$ and $z$ in ECEF coordinates is represented as
\begin{equation}\begin{split} g_x^m= \frac{(x_{\sf initial} -x^m)}{\sqrt{(x_{\sf initial}-x^m)^2+(y_{\sf initial}-y^m)^2+(z_{\sf initial}-z^m)^2}},\\ g_y^m= \frac{(y_{\sf initial} -y^m)}{ \sqrt{(x_{\sf initial}-x^m)^2+(y_{\sf initial}-y^m)^2+(z_{\sf initial}-z^m)^2}}, \\ g_z^m= \frac{(z_{\sf initial} -z^m)}{\sqrt{(x_{\sf initial}-x^m)^2+(y_{\sf initial}-y^m)^2+(z_{\sf initial}-z^m)^2}}\label{eq-8} \end{split} \end{equation}

For $M$ satellites, the linearized pseudorange measurements
\begin{equation} \begin{bmatrix} \delta\rho^1 \\ \delta \rho^2 \\ \vdots \\ \delta \rho^M \end{bmatrix} = \begin{bmatrix} (\mathbf{g}_{\sf initial}^1)^T \; 1 \\ (\mathbf{g}_{\sf initial}^2)^T \; 1 \\ \vdots \\ (\mathbf{g}_{\sf initial}^M)^T \; 1 \end{bmatrix}
\begin{bmatrix} \delta\mathbf{x} \\ \delta b_r \end{bmatrix}+\begin{bmatrix} \tilde{\epsilon}_{\rho}^1 \\ \tilde{\epsilon}_{\rho}^2\\ \vdots \\ \tilde{\epsilon}_{\rho}^{M} \end{bmatrix}, \label{eq5} \end{equation}

 where $\begin{bmatrix} (\mathbf{g}_{\sf initial}^1)^T \; 1 \\ (\mathbf{g}_{\sf initial}^2)^T \; 1 \\ \vdots \\ (\mathbf{g}_{\sf initial}^M)^T \; 1 \end{bmatrix}= \mathbf{G}$ is the geometry matrix with $M \times4$ dimensions which characterizes the relative geometry of a satellite-receiver.

\subsubsection{LOS vector}
\label{sec:los}
The LOS vector is crucial in determining the LOS direction between satellites and the receiver. {By encapsulating the} LOS vector along $x, y,$ and $z$ in ECEF coordinates, {it provides essential information for accurate positioning and navigation solutions. The LOS vector along $x, y,$ and $z$} is represented as \eqref{eq-8}.

\subsubsection{GDOP}
\label{sec:gdop}
The location geometry of the satellites with respect to the UT causes a change in the accuracy of position, which is coined as GDOP. The uncertainty in location measurements forms the region of uncertainty and ambiguity that is directly related to the relative geometry between the satellites in view and the receiver (UT). GDOP considers both three-dimensional region and timing error simultaneously.  
The multipath effect depends on the user-to-satellite geometry \cite{blanco2010satellite}. Moreover, in urban areas, it increases positional uncertainty and error.
A lower value of GDOP ensures less uncertainty and ambiguity in the determination of position. The GDOP is defined mathematically as follows:
\begin{equation} \sf{GDOP}= \sqrt{\left(\operatorname{tr}(\mathbf{G}^{T}\mathbf{G})^{-1}\right)} \label{eq6} \end{equation}

\subsubsection{$C⁄N_0$ and elevation angle}
\label{sec:ele}
 The NLOS signal is reflected or refracted to the GNSS receiver that causes signal attenuation. In most cases, the NLOS signals have smaller $C⁄N_0$ than LOS signals. Therefore, $C⁄N_0$ is an important indicator for detecting LOS and NLOS signals. Other than $C⁄N_0$, satellite elevation angle can also indicate the quality of pseudorange measurements. A lower elevation angle insists on a large NLOS position error; however, a large value indicates less multipath effect.

In general, the elevation angle affects GDOP and $C⁄N_0$. A favorable GDOP is achieved when satellites are at a low elevation angle. However, while low elevation angle can increase GDOP, it also reduces the $C⁄N_0$, especially in urban environments where multipath tends to affect the receiver's measurement \cite{blanco2011relation}. Moreover, the relationship between GDOP and elevation angle is non-linear in urban environments.

\subsubsection{Position correction}
\label{sec:pos}
Consider the ground truth position, $\mathbf{x}$ and the initial guess position, $\mathbf{x}_{\sf {initial}}$. The position correction ($\delta \mathbf{x}=\mathbf{x}-\mathbf{x}_{\sf {initial}}$) is mapped with pseudorange residuals, LOS vector, GDOP, elevation angle, and $C⁄N_0$. 

In this research, we consider all these as features along with the positioning error to train the \mbox{PC-DeepNet} position correction framework. The pseudorange measurements of satellites for positioning are a sequence of measurements that can be considered a set. Consider at a certain time ($t$) the features extracted from available satellite measurements set as
\begin{equation} X(t)=\{\mathbf{v}_1 (t),\mathbf{v}_2 (t),\mathbf{v}_3 (t) \hdots ,\mathbf{v}_M (t)\}, \label{eq7} \end{equation}
where $\mathbf{v}_i \in \mathbb {R}^d$, $d$-dimensional feature vectors, and $i$ is the index of satellite. In this paper, we consider seven (7) different features. Hence, $d=7$. { At a certain time ($t$) the feature vector of $i$-th satellite is given bellow,}
\begin{equation}
\mathbf{v}_i(t)=( \delta \rho^i,g_x^i,g_y^i,g_z^i,{\sf {GDOP}}, {\sf {Elevation}}^{i},{C⁄N_o}^i)
 \end{equation}

\subsection{Permutation-Invariant DNN Model}
\label{DNN_model}
Our objective is to map the position correction, $\delta \mathbf{x}(t)$ with the calculated feature vector $\mathbf{v}_i(t)$. If there are $M$ number of visible satellites, the position correction can be mapped by a function of $M$ features vectors,
 \begin{equation}
  \delta \mathbf{x}(t)=f\left(\mathbf{v}_1\left(t\right),\mathbf{v}_2 \left(t\right),\mathbf{v}_3 \left(t\right) \hdots,\mathbf{v}_M \left(t\right)\right)
 \end{equation}
 
The number and order of satellite measurements change frequently from the receiver side. Hence, traditional DNN models based on fixed dimensional vectors face the challenge of handling such a dynamic environment. Therefore, we propose \mbox{PC-DeepNet} that incorporates PI objective functions (sum pooling) on {input} sets where the output does not change when the input is reordered. This important characteristic of PI-DNN overcomes the limitation of the conventional DNN model that can handle only ordered inputs.
 The proposed \mbox{PC-DeepNet} is used to learn the functional mapping $\Psi$ that map the $M$ number of features vectors to the corresponding position correction. The function mapping $\Psi$ yields,
  \begin{equation}
 \delta \mathbf{x}(t)=\Psi \left(\mathbf{v}_i(t) \right), \forall i\in \{1,\hdots M\}
 \end{equation}

\begin{figure*}
    \centering
    \includegraphics[width=\linewidth]{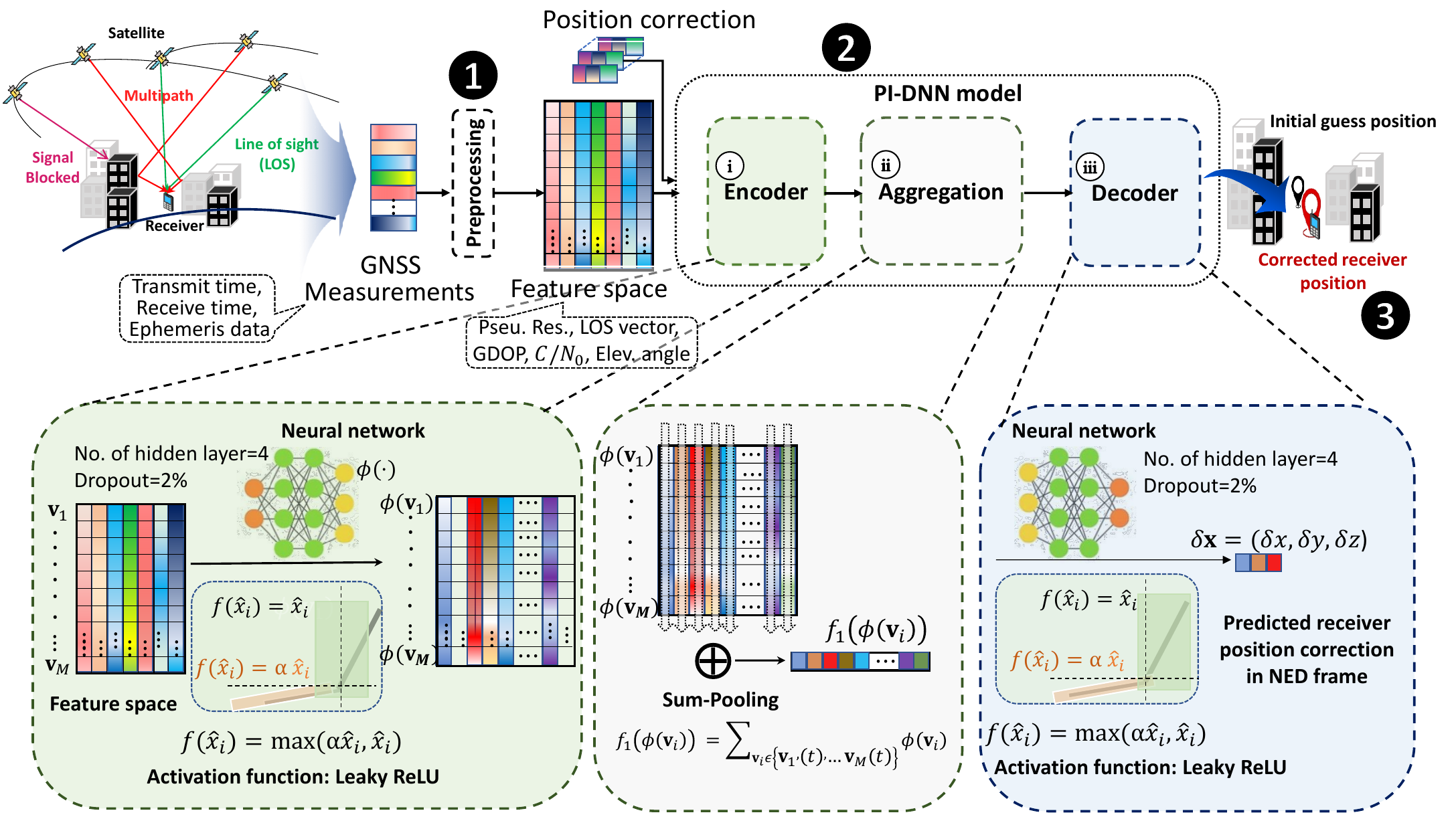}
    \caption{Architecture of PI-DNN model to minimize the positioning error in NLOS
and multipath environments.}
    \label{dnn_model}
\end{figure*}

Since our target domain consists of $M$ number of satellites, the number of possible inputs would be the power set of $ \chi=\num{2}^M$. 
Our proposed PI-DNN model consists of three sections: Encoder, Aggregation, and Decoder. Fig.~\ref{dnn_model} shows the PI-DNN model to minimize the positioning error in NLOS and multipath environments. This model aims to implement the following function on the measurement set $X(t)$.
\begin{equation} \begin{split}
\Psi \left(\mathbf{v}_i(t) \right)= \psi \left(\Sigma_{\mathbf{v}_i (t)\in {\left(\mathbf{v}_1 \left(t\right),\hdots,\mathbf{v}_M \left(t\right) \right)}} \phi(\mathbf{v}_i(t)) \right) 
\label{eq8} \end{split}, \end{equation}
where $i$ is the satellite index, $\mathbf{v}_i (t)$ is the row vector contains the feature, the encoder layer function $\phi(.)$, and the decoder layer function $\psi(.)$.

\subsubsection{Encoder}
\label{sec:encoder}
The encoder layer function $\phi(.)$ utilizes four fully connected layers of size $\{32,64,128,256\}$ with the Leaky ReLU activation function. {ReLU activation function does not activate all the neurons at the same time. The neurons will only be deactivated if the output of the linear transformation is less than 0. Since only a certain number of neurons are activated, the ReLU activation function is far more computationally efficient when compared to the other activation functions.} Unlike the ReLU activation function, Leaky ReLU provides the benefit of using the controlled negative portion of feature data, which is beneficial in our domain to map the features with position error.

Therefore, our proposed model uses Leaky ReLU with a setting $\alpha$ value to 0.1 (set value empirically). {Since Leaky ReLU ensures that all the neurons in the network contribute to the output, overfitting may arise.} To address the overfitting issue, $2\%$ dropout is applied in the third layer.

\subsubsection{Aggregation}
\label{sec:Aggregation}
The aggregation layer aggregates the output of the encoder using a PI {objective} function (sum-pooling) \cite{wagstaff2022universal} as
\begin{equation}
f_1\left(\phi(\mathbf{v}_i)\right) =\Sigma_{\mathbf{v}_i \in {\left(\mathbf{v}_1 \left(t\right),\hdots,\mathbf{v}_M \left(t\right) \right)}} \phi(\mathbf{v}_i) 
\label{eq9}\end{equation}
\subsubsection{Decoder}
\label{sec:decoder}
Finally, the decoder layer $\psi(.)$ that uses four fully-connected layers of size $\{\num{256},\num{128},\num{64},\num{32}\}$ with Leaky ReLU activation function to predict the position correction along $x, y,$ and $z$ direction. A $2\%$ dropout layer is applied to prevent overfitting.

The training set includes extracted features ($X$) with corresponding position correction as target $(\delta \mathbf{x})$. The training set is used to select the hyperparameters of the proposed model, and a validation set is used to evaluate its performance. The proposed PI-DNN model is trained for up to $\num{100}$ epochs with batch sizes of $\num{8}$, considering mean-squared error (MSE) as a loss function. 
Compared to traditional stochastic gradient descent (SGD) \cite{bottou2012stochastic}, Adam optimizer \cite{kingma2014adam} maintains separate adaptive learning rates for each parameter, which can be particularly advantageous in our interested scenarios where the features have varying scales \cite{ruder2016overview}. This adaptability helps the model to converge efficiently and avoid issues related to setting a global learning rate for all parameters.
This model utilizes Adam optimizer to minimize the error by setting parameters $\alpha_l=0.001$ (learning rate), $\beta_1= 0.9$ (decay rate for the first moment), $\beta_2=0.999$ (decay rate for the second moment) and $\epsilon=10^{-8}$ (constant to sum of mini-batch variances).

\subsection{Corrected Receiver Position}
\label{cor_rec}

The initial guess position $\mathbf{x}_{\sf initial}(t)$ is obtained using pseudorange-based r-WLS solution \cite{robust_WLS} to reduce the influence of outliers on the solution using smooth L1 loss function. We get the input feature vector $\mathbf{v}_i(t)$ following \ref{sec: Proposed_methods} and the initial guess position using \ref{sec: preprocess}. The proposed PI-DNN model predicts the position correction $\delta\mathbf{x}(t)$ using $\mathbf{v}_i(t)$ as input. Then the position correction is added with the initial guess position, $\mathbf{x}_{\sf initial}(t)$ to get the final corrected position, $\mathbf{\hat{x}}(t)$.
 \begin{equation}
 \label{eq10}
     \mathbf{\hat{x}}(t)=\mathbf{x}_{\sf initial}(t)+\delta \mathbf{x}(t)
 \end{equation}

\section{Dataset Description}
\label{sec: Dataset}
Google released two datasets collected from multiple Android phones, including raw GNSS measurement and ground truth \cite{fu2020android,fu2022android}
with less geographic diversity data and are publicly available to use.
These datasets are collected from multiple driving paths in the urban and sub-urban public areas in San Francisco Bay and Los Angeles. This paper refers to two datasets, dataset-I and dataset-II, published in 2021 and 2022, respectively. The dataset-I has $\num{29}$ training traces (driving trajectory) collected from multiple Android phones (Google Pixel4, Google Pixel 4XL, SamsungS20Ultra, and Xiaomi Mi8). The dataset-I contains $C/N_0$, elevation angle, Doppler rate, satellite transmit time, signal arrival time, and other raw GNSS measurements of L1, L5 channels from GPS, Galileo, GLONASS, Beidou, and QZSS. NovAtel SPAN system is utilized to get precise ground truth location files. On the other hand, dataset-II published more datasets spanning more routes and containing $\num{62}$ traces, including all the traces in dataset-I. Like dataset-I, dataset-II also follows the same data collection method. 
To ensure effective training of the DNN model, a large dataset is required. Given the extensive data involved, we employ offline training for the proposed \mbox{PC-DeepNet} framework. After training, position correction can be estimated using this model. As a result, the long training time will not affect the device performance while determining the corrected position using the PI-DNN model.

\begin{figure}[t]
    \centering
    \includegraphics[width=\linewidth]{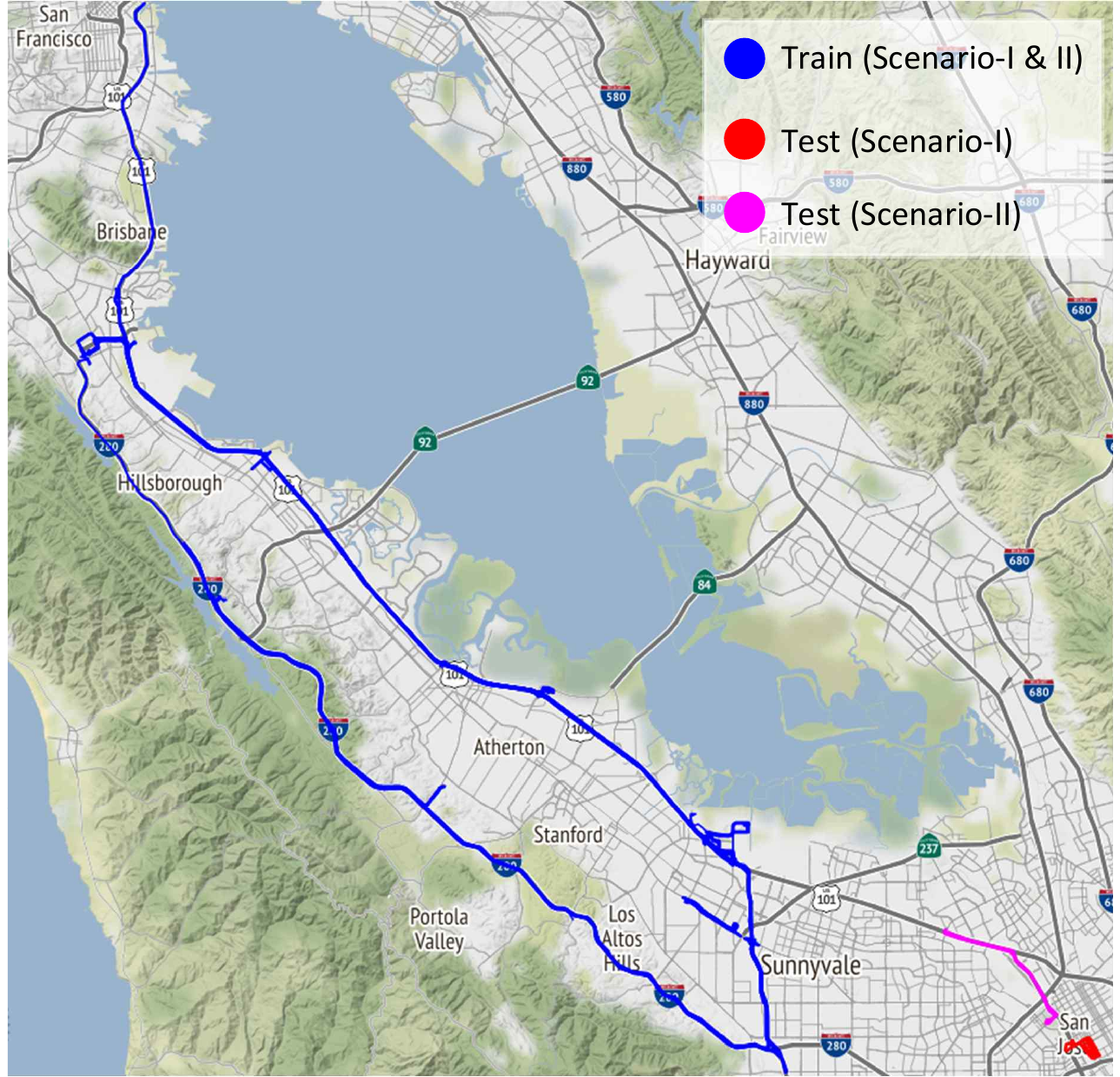}
    \caption{ Test and train location map in San Francisco.}
    \label{map-1}
\end{figure}

\section{Result and Discussion}
\label{sec: Result_and_Discussion}

This section presents the findings derived in the case of the proposed \mbox{PC-DeepNet} framework. Firstly, we present the dataset description for all three scenarios that are considered to evaluate the proposed framework. Then, we include the positioning outcomes with a state-of-the-art method pseudorange-based r-WLS and the proposed framework for all three scenarios. After that, the 2D and 3D-positioning error performance evaluation and comparison with state-of-the-art methods (WLS\cite{morton2021position}, r-WLS\cite{robust_WLS}, and KF\cite{sever2022gnss}) results are explained. In addition, the positioning error and computational complexity of the proposed methods are compared with the ML-based approach in \cite{kanhere2022improving}.

\begin{figure}[t]
    \centering
    
    \includegraphics[width=\linewidth]{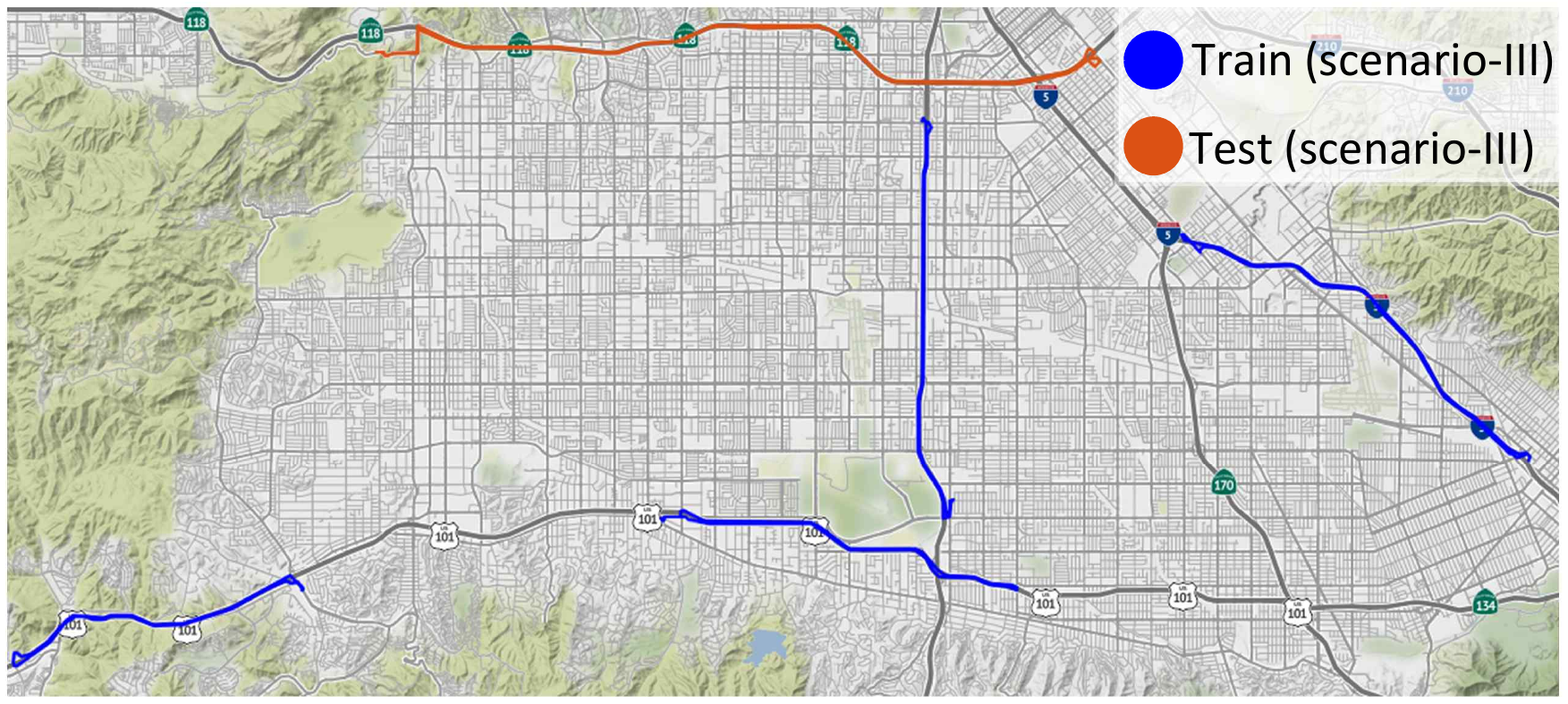}
    \caption{ Test and train location map in Los Angeles.}
    \label{map-2}
\end{figure}

\begin{figure}[t]
    \centering
    \includegraphics[width=.8\linewidth]{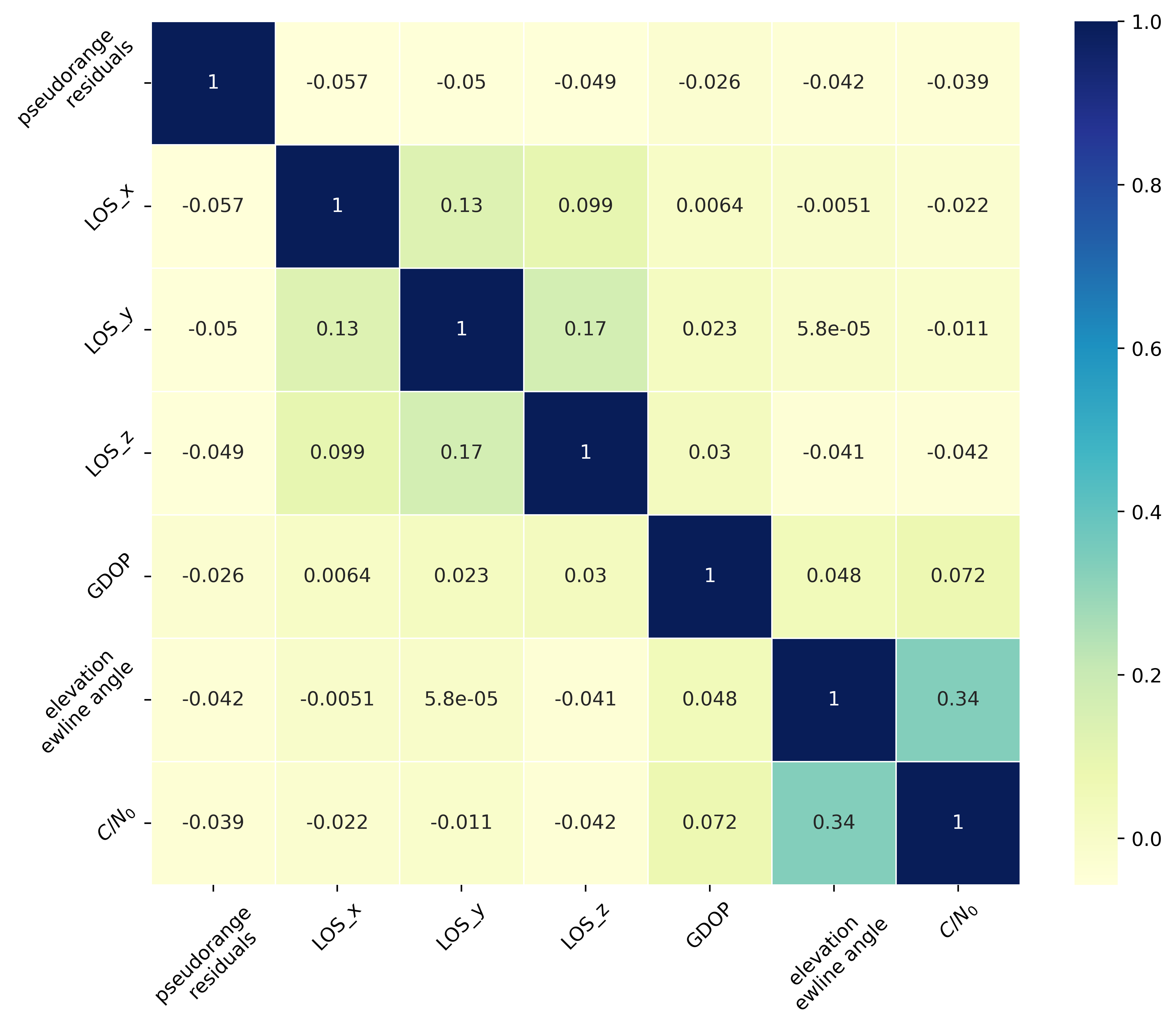}    
    \caption{Kendall's tau correlation among different features including pseudorange residuals, LOS vector, GDOP, elevation angle, and $C/N_0$.}
    \label{tau}
\end{figure}

We consider three different scenarios to evaluate the proposed framework. Among the three scenarios, scenario-I and scenario-II are selected in urban and sub-urban areas in San Francisco, and scenario-III in urban areas in Los Angeles. {We use a training-validation-test split of $\num{75}\%$, $\num{10}\%$, and $\num{15}\%$ of data. In scenario-I, fifty-five different driving trajectories consisting of $\num{95542}$ data samples in dataset-I are used for training, validation, and testing in the San Francisco area. Fifty-five different driving trajectories consisting of $\num{94995}$ in dataset-II are used for training, validation, and testing in scenario-II in the San Francisco area. Fig.~\ref{map-1} shows the map used for scenario-I and scenario-II. Blue marked trajectory indicates the training location, whereas red and purple marked trajectories show the testing locations for scenario-I and scenario-II, respectively. In scenario-III, six different traces of $\num{16047}$ data samples are used for training, validation, and testing in dataset-II of Los Angeles. For all the cases, the training and test traces are different.} Fig.~\ref{map-2} shows the map of scenario-III where train and test locations are marked with blue and orange colors.
To illustrate the non-linear correlation among different features, we calculate Kendall's tau correlation \cite{essam2022comparison} among different features. Fig.~\ref{tau} shows Kendall's tau correlation among different features (pseudorange residuals, LOS vector, GDOP, elevation angle, and $C/N_0$). From the figure, it is obvious that there is no strong correlation among different features except $0.34$ between the elevation angle and $C/N_0$, which is moderated.
Therefore, we consider those features to map with the position error using suitable DNN architectures.

\begin{figure*}[h]
    \centering
    \includegraphics[width=.8\linewidth]{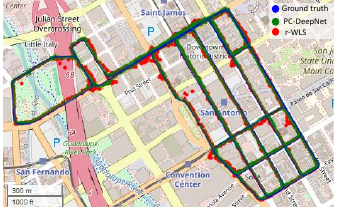}
    \caption{ Evaluation map for urban area in San Francisco.}
    \label{map-3}
\end{figure*}

\begin{figure*}[h]
    \centering
    \includegraphics[width=.8\linewidth]{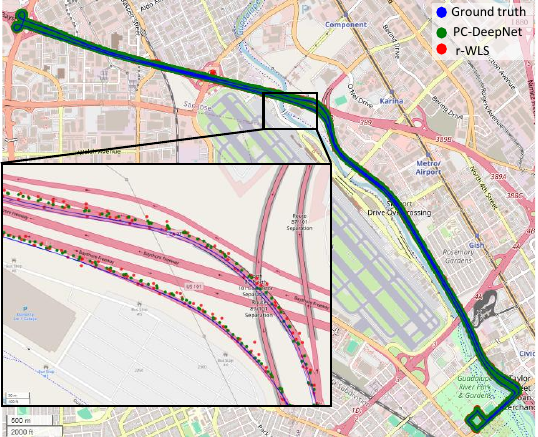}
    \caption{ Evaluation map for the sub-urban area in San Francisco.}
    \label{map-4}
\end{figure*}

\begin{figure*}[h]
    \centering
    \includegraphics[width=.8\linewidth]{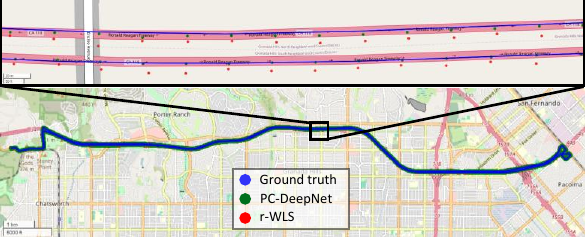}
    \caption{ Evaluation map for urban area in Los Angeles.}
    \label{map-5}
\end{figure*}

\begin{figure}[h]
    \centering
    \includegraphics[width=0.6\linewidth]{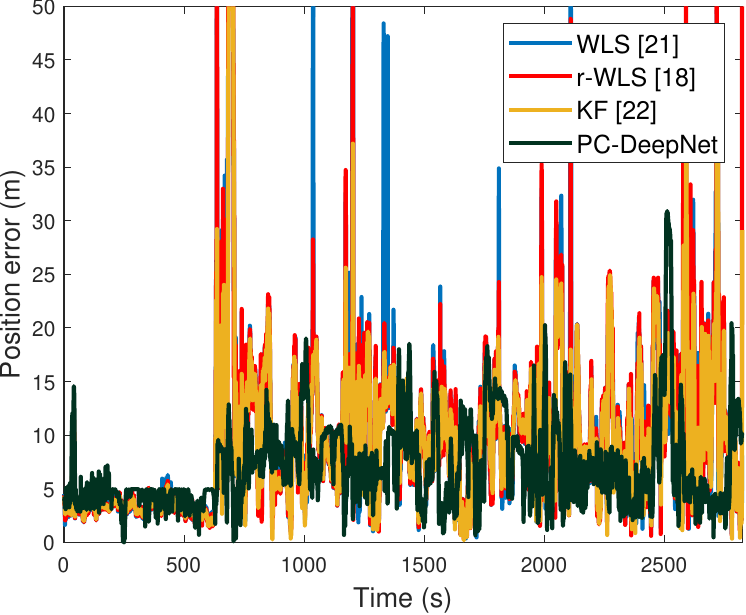}
    \caption{{Time series comparison of error between different methods in scenario-I.}}
    \label{time-s1}
\end{figure}

\begin{figure}[h]
    \centering
    \includegraphics[width=0.6\linewidth]{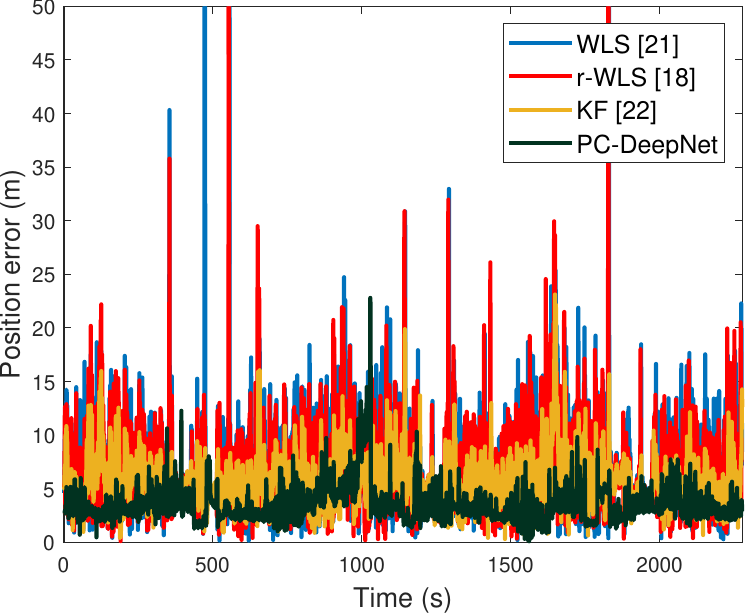}
    \caption{Time series comparison of error between different methods in scenario-II.}
    \label{time-s2}
\end{figure}

\begin{figure}[h]
    \centering
    \includegraphics[width=0.6\linewidth]{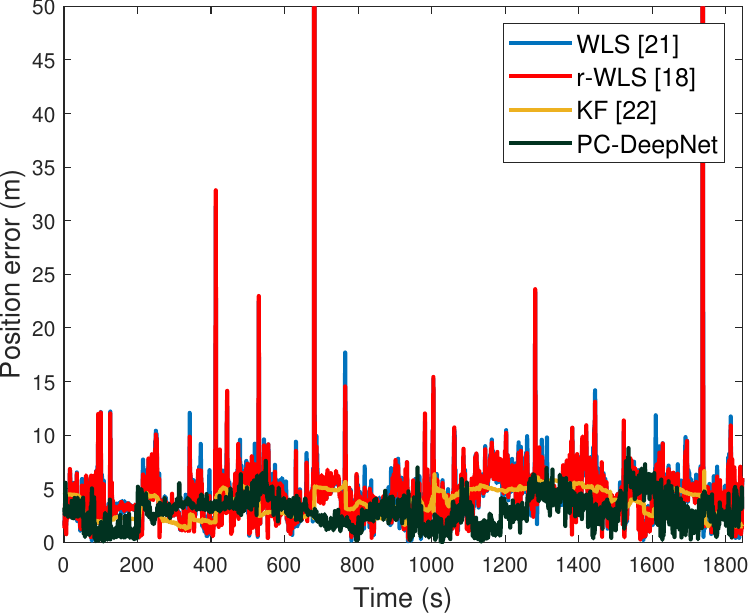}
    \caption{Time series comparison of error between different methods in scenario-III.}
    \label{time-s3}
\end{figure}

The positioning outcomes with a state-of-the-art method, pseudorange-based r-WLS and the proposed framework (\mbox{PC-DeepNet}) for scenario-I, scenario-II, and scenario-III shown in Figs.~\ref{map-3},~\ref{map-4}, and~\ref{map-5}, respectively. The ground truth, r-WLS, and proposed framework (\mbox{PC-DeepNet}) are marked with blue, red, and green colors.

The horizontal positioning (latitude, longitude) errors in each estimation are evaluated by a score calculated using the mean of the $50$-th percentile and $95$-th percentile of the horizontal positioning errors. Firstly, the corrected position in ECEF coordinates is converted to geodetic coordinates (latitude, longitude, and altitude). Then the converted latitude $\Phi_n$, longitude $\Lambda_n$, and corresponding horizontal error $(D_n)$ for $n$-th trace is defined as
\begin{eqnarray}
\label{eq11}
\Phi^n =\{\phi^n_1,\phi^n_2,\cdots\,\phi^n_k\},\nonumber\\
\Lambda^n =\{\lambda^n_1,\lambda^n_2,\cdots\,\lambda^n_k\},\nonumber\\
D^n =\{d^n_1,d^n_2,\cdots\,d^n_k\},
\end{eqnarray} 
where $k$ refers to the total number of epochs, $\lambda$ is latitude, and $\phi$ is altitude in each trace.

We use Vincenty distance \cite{vincenty1975geodetic} to calculate the horizontal positioning (latitude, longitude) error between two pairs of estimated and ground truth positions, respectively. Vincenty method deals with an iterative approach that assumes Earth as an oblate spheroid other than a perfect sphere and provides a higher degree of accuracy than Haversine method \cite{goodwin1910haversine}. The horizontal error $(d_j)$ at each epoch $j$ is defined using Vincenty distance as follows
\begin{equation}
 d_j = \mathrm {vincenty} \left( \left(\phi_j,\lambda_j\right),\left(\phi_{\mathsf{gt}_{j}},\lambda_{\mathsf{gt}_j}\right) \right),  
\end{equation}
where $\mathsf{gt}$ refers to the ground truth of the epoch in each trace.

The time series positioning error analysis presents the effectiveness of the methods over time. 
We can clearly guess the error magnitude of the positioning method over time and thus 
evaluate the method easily.
We provide time series positioning of error performance comparison of the proposed framework with existing state-of-the-art methods in three different scenarios (scenario-I, scenario-II, and scenario-III) as shown in Figs. \ref{time-s1}, \ref{time-s2}, and \ref{time-s3}. The positioning method utilizing pseudorange-based WLS gets a 2D-position error that varies from $0.35$ m to $129.37$ m with a mean absolute error (MAE) of $10.03$ m in scenario-I. The positioning method utilizing pseudorange-based r-WLS gets a 2D-position error that varies from $0.29$ m to $87.91$ m with MAE of $8.81$ m in scenario-I. The positioning method utilizing pseudorange-based KF gets a 2D-position error that varies from $0.18$ m to $67.53$ m with an MAE of $8.39$ m in scenario-I. Our proposed \mbox{PC-DeepNet} gets 2D-position error that varies from $0.08$ m to $30.90$ m with MAE of $6.95$ m in scenario-II. 
The WLS-based positioning method gets a 2D-position error that varies from $0.30$ m to $116.98$ m with MAE of $6.65$ m in scenario-II. In the case of r-WLS, the positioning error varies from $0.26$ m to $68.87$ m with an MAE of $6.38$ m in scenario-I. The KF-based method gets a 2D-position error ranging from $0.23$ m to $23.11$ m with an MAE of $5.64$ m in scenario-II. Our proposed \mbox{PC-DeepNet} gets a 2D-position error ranging from $0.16$ m to $22.80$ m with an MAE of $3.27$ m. 
Moreover, in scenario-III, the WLS-based positioning method gets a 2D-position error ranging from $0.31$ m to $173.33$ m with an MAE of $4.08$ m. In the case of r-WLS, the positioning error varies from $0.27$ m to $173.33$ m with an MAE of $3.97$ m. The KF-based method gets a 2D-position error that varies from $0.27$ m to $6.63$ m with an MAE of $3.97$ m. Our proposed \mbox{PC-DeepNet} gets 2D-position error ranging from $0.13$ m to $8.78$ m with an MAE of $2.65$ m.
From all the time series figures, it is obvious that our proposed \mbox{PC-DeepNet} performs better than other methods WLS \cite{morton2021position}, r-WLS \cite{robust_WLS}, and KF \cite{sever2022gnss}.

\begin{figure}[h]
    \centering
    \includegraphics[width=0.7\linewidth]{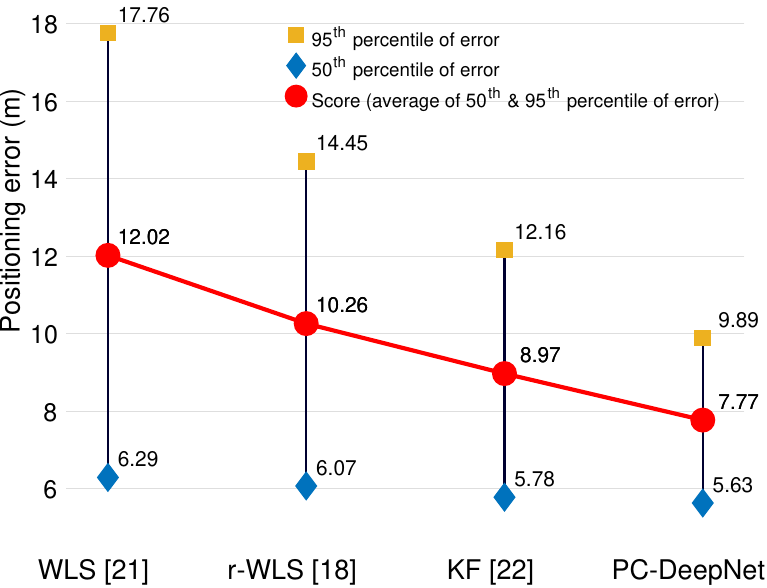}
    \caption{{Comparison of horizontal positioning error among state-of-the-art methods with the proposed framework (PC-DeepNet).}}
    \label{score}
\end{figure}

\begin{table}[h]\renewcommand{\arraystretch}{1}
\centering
\caption{Performance evaluation of positioning error using Vincenty distance.}
\begin{tabular}{|l|c|c|r|}
\hlineB{3}\small
\textbf{Scenario} & \textbf{Score (m)} & \textbf{50-th percentile} & \textbf{95-th percentile} \\
& & \textbf{of error (m)} & \textbf{of error (m)} \\
\hlineB{3}
Scenario-I & $11.98$ & $7.86$ & $16.09$\\
 Scenario-II & $6.76$ & $5.39$ & $8.12$\\
 Scenario-III & $4.56$ & $3.64$ & $5.48$\\
 
  \hlineB{3}
\end{tabular}
  \label{tab-1}
\end{table}

The performance evaluation of 2D-positioning error for the proposed \mbox{PC-DeepNet} framework in different scenarios along horizontal (latitude and longitude) direction using Vincenty distance is tabulated in Table~\ref{tab-1}. The score is the mean of $50$-th percentile and $95$-th percentile of the positioning error. 
Considering $N$ number of total traces, the score {\cite{kanhere2022improving}} is defined mathematically as
\begin{equation}
 \mathrm{score} = \frac{1}{N} \sum_{n=1}^{N} \frac{\mathrm{percentile}({D_n},50)+\mathrm{percentile}({D_n},95)}{2},   
\end{equation}
where $\mathrm{percentile}(\textit{D},\textit{b})$ refers to the value in $D$ at which the $b$ percentage of the data lies in ascending order below that value.
In dense urban areas, there is a high probability of NLOS and multipath signals. Therefore, the score is large ($11.98$ m) in scenario-I due to dense urban areas compared to scenario-II and scenario-III, which are sub-urban areas having scores of $6.76$ m and $4.56$ m, respectively.

The \mbox{2D} positioning error with Vincenty distance  
of the proposed framework (\mbox{PC-DeepNet}) is compared with state-of-the-art methods WLS \cite{morton2021position}, r-WLS \cite{robust_WLS}, and KF \cite{sever2022gnss}. Fig.~\ref{score} shows the comparison of 2D-positioning error for WLS, r-WLS, {KF,} and the proposed \mbox{PC-DeepNet}. 
The vertical (altitude) error is not considered here. The horizontal (latitude and longitude) error is only used for evaluation. The proposed framework achieves less distance error than the other methods. 
The proposed \mbox{PC-DeepNet} gets a score value of $7.77$ m, which is less than WLS ($12.02$ m){,} r-WLS ($10.26$ m){, and KF ($8.97$ m)} approaches.

\begin{table}[h]\renewcommand{\arraystretch}{1}
\centering
\caption{Performance evaluation of positioning error along NED coordinates.}
\begin{tabular}{|l|c|c|r|}
\hlineB{3}
\textbf{Scenario} & \textbf{North (m)} & \textbf{East (m)} & \textbf{Down (m)} \\
\hlineB{3}
Scenario-I & $4.73\pm2.95$ & $5.10\pm4.88$ & $5.24\pm4.69$\\
 Scenario-II & $2.14\pm1.91$ & $2.48\pm1.14$ & $6.04\pm5.72$\\
 Scenario-III & $2.32\pm1.61$ & $1.29\pm1.01$ & $4.44\pm3.70$\\
 
  \hlineB{3}
  \end{tabular}
  \label{tab-2}
\end{table}

Since the Vincenty method considers horizontal position (latitude and longitude) error, to get an intuition of 3D-position error, the position error using North, East, Down (NED) coordinates is calculated all the three scenarios. Table~\ref{tab-2} tabulated the performance evaluation of positioning error along the NED coordinates. The 3D-position errors are $8.70$ m, $6.87$ m, and $5.17$ m for scenario-I, scenario-II, and scenario-III, respectively. These results clearly indicate that the positioning error is more pronounced in denser urban environments, as exemplified by scenario-I. 

\begin{figure}[t]
    \centering
    \includegraphics[width=.7\linewidth]{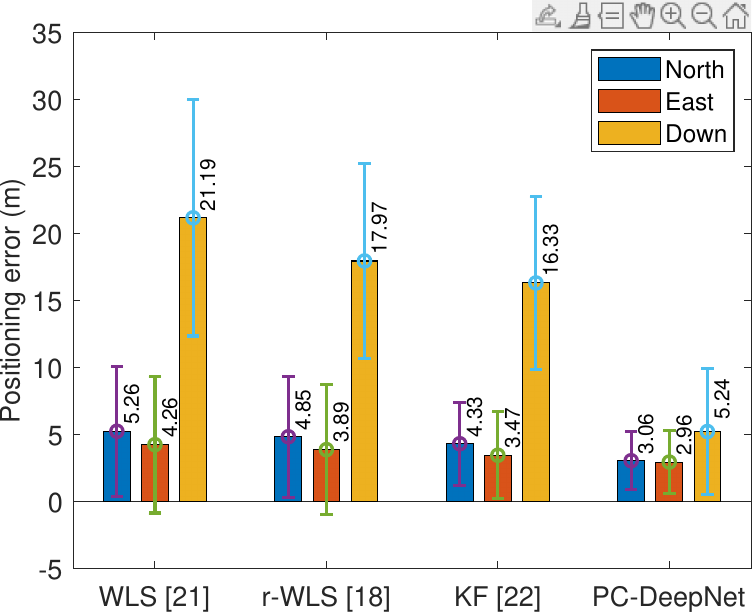}
    \caption{Performance comparison of positioning MAE among state-of-the-art methods with the proposed framework (PC-DeepNet) in NED coordinates.}
    \label{NED-error}
\end{figure}

Fig. \ref{NED-error} shows the MAE in the NED coordinates. {Three} different methods: WLS{,} r-WLS, {and KF} are considered with the proposed \mbox{PC-DeepNet} for comparison of positioning error in all three scenarios. 
The average positioning MAE for all three scenarios in terms of mean and standard deviation in the north direction for WLS, r-WLS, {KF,} and the proposed \mbox{PC-DeepNet} are $5.26\pm4.84$ m, $4.85\pm4.53$ m, {$4.33\pm3.09$ m,} and $3.06\pm2.16$ m, respectively. The average positioning MAE for all three scenarios in terms of mean and standard deviation in the east direction for WLS, r-WLS, {KF,} and the proposed \mbox{PC-DeepNet} are $4.26\pm5.10$ m, $3.89\pm4.83$ m, {$3.47\pm3.24$ m,} and $2.96\pm2.34$ m, respectively. Moreover, the average positioning MAE for all three scenarios in terms of mean and standard deviation in the down direction for WLS, r-WLS, {KF,} and the proposed \mbox{PC-DeepNet} are $21.19\pm8.82$ m, $17.97\pm7.29$ m, {$16.33\pm6.48$ m,} and $5.24\pm4.72$ m, respectively.

\begin{table}[H]\renewcommand{\arraystretch}{1}
\centering
\caption{Performance evaluation of positioning error along NED coordinates with confidence interval.}
\begin{tabular}{|l|c|c|r|}
\hlineB{3}
& \multicolumn{3}{c|}{$95$\% Confidence Interval}\\ \cline{2-4}
\multirow{-2}{*}{\textbf{Scenario}} & \textbf{North (m)} & \textbf{East (m)} & \textbf{Down (m)} \\
\hlineB{3}
Scenario-I & $4.62\sim4.85$ & $4.86\sim5.25$ & $4.88\sim5.47$\\
 Scenario-II & $2.02\sim2.19$ & $2.41\sim2.51$ & $5.71\sim6.25$\\
 Scenario-III & $2.23\sim2.39$ & $1.23\sim1.33$ & $4.15\sim4.61$\\
 
  \hlineB{3}
  \end{tabular}
  \label{tab-2_95}
\end{table}

The $95\%$ confidence interval is a range of values that you can be $95\%$ confident contains the true mean of the population. we have calculated the $95\%$ confidence interval for the positioning error along NED coordinates and tabulated in Table~\ref{tab-2_95}. The performance evaluation of positioning error along NED coordinates presented in Table~\ref{tab-2_95} shows that the estimated position fall into $95\%$ confidence interval.

Kanhere \textit{et al.} \cite{kanhere2022improving} consider the initial position guess utilizing randomly sampled noise with the magnitude of $\eta$ ($15$ m and $30$ m) to the true position while calculating position error for training the DNN model. The model achieved maximum accuracy in the case of $\eta\!=\!15$ m. They utilized only dataset-I for the evaluation of their DNN model. Whereas we used the position outcome of r-WLS as the initial guess and calculated the position error with respect to it. To compare the outcome of Kanhere \textit{et al.} \cite{kanhere2022improving} with our proposed framework, we also used dataset-I. {The positioning MAE of the NED positioning error of different methods for scenario-I is} tabulated in Table ~\ref{tab-3}. {The positioning MAE in terms of mean and standard deviation in the north direction for WLS, r-WLS, KF, Kanhere \textit{et al.}, and the proposed \mbox{PC-DeepNet} are $6.49\pm5.68$ m, $5.59\pm4.93$ m, $5.29\pm5.29$ m, $6.4\pm5.2$ m, and $4.75\pm2.95$ m, respectively. The positioning MAE in terms of mean and standard deviation in the east direction for WLS, r-WLS, KF, Kanhere \textit{et al.}, and the proposed \mbox{PC-DeepNet} are $7.65\pm8.49$ m, $6.81\pm7.80$ m, $6.52\pm7.13$ m,$5.9\pm5.0$ m, and $5.10\pm4.86$ m, respectively. Moreover, the positioning MAE in terms of mean and standard deviation in the down direction for WLS, r-WLS, KF, Kanhere \textit{et al.}, and the proposed \mbox{PC-DeepNet} are $56\pm16.35$ m, $46.49\pm12.15$ m, {$39.8.33\pm9.62$ m,}$6.2\pm4.9$ m, and $5.26.2\pm4.9$ m, respectively.} Table ~\ref{tab-3} shows that {the} proposed framework performs better than the other.

\begin{table}[ht]\renewcommand{\arraystretch}{1}
    \centering
    \caption{Performance evaluation of positioning error for scenario-I.}
    \begin{tabular}{|l|c|c|r|}
    \hlineB{3}
     \textbf{Method} & \textbf{North (m)} & \textbf{East (m)} & \textbf{Down (m)} \\
    \hlineB{3}
    WLS \cite{morton2021position} & $6.49\pm5.68$ & $7.65\pm8.49$ & $56\pm16.35$ \\
    r-WLS \cite{robust_WLS} & $5.59\pm4.93$ & $6.81\pm7.80$ & $46.49\pm12.15$ \\
    KF \cite{sever2022gnss} & $5.29\pm5.29$ & $6.52\pm7.13$ & $39.8\pm9.62$ \\
     Kanhere \textit{et al.} \cite{kanhere2022improving} & & & \\
   with $\eta=15$ m &  \multirow{-2}{*}{$6.4\pm5.2$} & \multirow{-2}{*}{$5.9\pm5.0$} & \multirow{-2}{*}{$6.2\pm4.9$}\\
      \hline
 PC-DeepNet & $\textbf{4.73}\pm\textbf{2.95}$ & $\textbf{5.10}\pm\textbf{4.88}$ & $\textbf{5.24}\pm\textbf{4.69}$\\
    \hlineB{3}
    \end{tabular}
    \label{tab-3}
\end{table}

\begin{table}[h]\renewcommand{\arraystretch}{1.2}
    \centering
    \caption{{Complexity analysis of the proposed PI-DNN model.}}
    \label{tab:parameter}
    \begin{tabular}{|c|l|r|}
    \hlineB{3}
        \textbf{Section}&\textbf{Layer type} & \textbf{\# Parameters} \\
        \hlineB{3}
        &Linear-1 & \num{256} \\ 
        &LeakyRelu & \num{0} \\ 
        &Linear-2 & \num{2112} \\ 
        &LeakyRelu & \num{0} \\ 
        &Linear-3 & \num{8320} \\ 
        &Dropout & \num{0} \\ 
        &LeakyRelu & \num{0} \\ 
        \multirow{-8}{*}{Encoder}&Linear-4 & \num{33024} \\ \hline
        Aggregation& Sum-pooling & \num{0}\\   \hline
        &Linear-5 & \num{32896} \\ 
        & LeakyRelu & \num{0} \\  
        &Linear-6 & \num{8256} \\ 
        &  LeakyRelu & \num{0} \\ 
        &Linear-7 & \num{2080} \\ 
        &  Dropout-15 & \num{0} \\ 
        & LeakyRelu & \num{0} \\ 
        & Linear-8 & \num{1056} \\ 
        & LeakyRelu & \num{0} \\ 
        \multirow{-10}{*}{Decoder}& Output & \num{33} \\  \hlineB{3}
         \multicolumn{3}{|r|}{Total parameters: \num{88033}} \\
        \hlineB{3}
    \end{tabular}
\end{table}

We have performed the complexity analysis for our proposed PI-DNN model. The proposed PI-DNN model comprises three modules: Encoder, Aggregation, and Decoder. The encoder and decoder utilize four linear layers with the Leaky ReLU activation function and dropout. The number of parameters in each layer is tabulated in Table~\ref{tab:parameter}.
In addition, we have also performed a comparison with Kanhere \textit{et al.} \cite{kanhere2022improving} in terms of the number of parameters and estimated memory size in kilobytes.
The number of parameters used in Kanhere \textit{et al.} \cite{kanhere2022improving} is $\num{151107}$, and the estimated memory is $611$ KB. For our proposed \mbox{PC-DeepNet}, the number of parameters and estimated memory size is $\num{88033}$ and $340$ KB, respectively, which is less than the Kanhere \textit{et al.} \cite{kanhere2022improving}.
The comparison result proves the superiority of the proposed approach over other methods. The new Samsung Galaxy S24 smartphone includes a GPU that makes it possible to infer with a less complex model.
The complexity of the model should be kept as low as possible to satisfy the hardware resources (memory and processing speed) in practical applications.

\begin{table}[ht]
    \centering
    \caption{{The effect of the number of layers in positioning error.}}
    \begin{tabular}{|c|c|l|r|}
    \hlineB{3} 
    \textbf{Model} & \textbf{\# Parameter} & \textbf{Scenario} & \textbf{3D pos. error (m)} \\
    \hlineB{3}
    Model-1 & & Scenario-I & $10.67$  \\  
    (2 linear-layer in encoder, & & Scenario-II & $9.05$  \\ 
      2 linear-layer in decoder) & \multirow{-3}{*}{\num{8641}}& Scenario-III & $7.99$ \\ 
     \hline
    Model-2 & & Scenario-I & $9.37$  \\  
     (3 linear-layer in encoder, & & Scenario-II & $8.41$  \\ 
       3 linear-layer in decoder) & \multirow{-3}{*}{\num{37569}}& Scenario-III & $7.72$ \\ 
     \hline
      \textbf{Proposed PI-DNN} & & Scenario-I & $\textbf{8.70}$  \\  
     (4 linear-layer in encoder, & & Scenario-II & $\textbf{6.87}$  \\ 
      4 linear-layer in decoder)  & \multirow{-3}{*}{$\textbf{88,033}$}& Scenario-III & $\textbf{5.17}$ \\
     \hline
     Model-3 & & Scenario-I & $10.67$  \\  
     (5 linear-layer in encoder, & & Scenario-II & $11.76$  \\ 
      5 linear-layer in decoder)  & \multirow{-3}{*}{\num{612545}}& Scenario-III & $9.79$ \\ 
     \hline 
    Model-4 & & Scenario-I & $10.48$  \\  
    (6 linear-layer in encoder, & & Scenario-II & $9.23$  \\ 
      6 linear-layer in decoder)  & \multirow{-3}{*}{\num{1137857}}& Scenario-III & $8.74$ \\  
    \hlineB{3}
    \end{tabular}
    \label{tab:model_layer_3}
\end{table}

We have considered four modified models of PI-DNN: model-1, model-2, model-3, and model-4, with different numbers of layers. We have performed a performance comparison with all the models, including the proposed PI-DNN model, and tabulated the 3D-position error in meters in Table~\ref{tab:model_layer_3}. Moreover, the number of parameters in each model is also reported in Table~\ref{tab:model_layer_3}. In the case of model-1, model-2, and the proposed PI-DNN model, while the number of layers in the model increases, the positioning error also decreases. For models 3 and 4, although the number of layers increases, the 3D-position error does not decrease. Our findings suggested that, beyond a certain point, adding more layers increased the susceptibility to overfitting \cite{ruder2016overview}.
In the design of our proposed PI-DNN model, we carefully consider the balance between model complexity and the risk of overfitting. Our current architecture strikes a balance between capturing complex relationships and generalization ability, as evidenced by our experimental results on the validation set.
From the evaluation result, it is obvious that the proposed framework minimizes the positioning error remarkably. Therefore, the proposed \mbox{PC-DeepNet} outperforms the other approaches regarding accuracy.

\section{Conclusion}
\label{sec: Conclusion}
This paper proposed \mbox{PC-DeepNet}, a deep neural network framework based on permutation-invariant architectures, designed to reduce the GNSS positioning error in IoT applications, particularly in urban and suburban environments. 
The model leverages features such as pseudorange residuals, LOS vectors, carrier-to-noise-ratio, elevation angle, and GDOP to construct set-based inputs, which are mapped to 3D-positioning errors. Validated using high-accuracy Android GNSS datasets collected in San Franciso Bay and Los Angeles, PC-DeepNet demonstrated superior performance in mitigating NLOS and multipath-induced errors compared to state-of-the-art methods. With minimal parameter requirements and a low memory footprint, it is well-suited for resource-constrained IoT devices. This model is trained using real-world correction data to capture urban-specific error patterns. Although the current implementation relies solely on GPS signals, integrating other constellations (Galileo, GLONASS, BeioDou, and QZSS) via a satellite selection algorithm will further enhance accuracy. Future work will include a broader geographic validation using ground-truth collected with high-precision hardware.

\bibliographystyle{IEEEtran}
 \bibliography{ref}
\end{document}